\documentclass[conference]{IEEEtran}
\ifCLASSINFOpdf
\else
\fi
\usepackage[utf8]{inputenc}
\usepackage{graphicx}
\usepackage{cite}

\begin{document}
\title{An electronic-game framework for evaluating coevolutionary algorithms}

\author{\IEEEauthorblockN{Karine da Silva Miras de Araújo}
\IEEEauthorblockA{Center of Mathematics, Computer e Cognition (CMCC)\\
	Federal University of ABC (UFABC)\\
	Santo André, Brazil \\
	E-mail: karine.smiras@gmail.com}
\and
\IEEEauthorblockN{Fabrício Olivetti de França}
\IEEEauthorblockA{Center of Mathematics, Computer e Cognition (CMCC)\\
		Federal University of ABC (UFABC)\\
		Santo André, Brazil \\
		E-mail: folivetti@ufabc.edu.br}}


%Telephone: (800) 555--1212, Fax: (888) 555--1212}


\maketitle

\begin{abstract}

One of the common artificial intelligence applications in electronic games consists of making an artificial agent learn how to execute some determined task
successfully in a game environment. One way to perform this task is through
machine learning algorithms capable of learning the sequence of actions required to win in a given game environment. There are several supervised learning
techniques able to learn the correct answer for a problem through
examples. However, when learning how to play electronic games, the correct
answer might only be known by the end of the game, after all the actions were
already taken. Thus, not being possible to measure the accuracy of each
individual action to be taken at each time step. A way for dealing with this
problem is through Neuroevolution, a method which trains Artificial Neural Networks using
evolutionary algorithms. In this article, we introduce a framework for testing
optimization algorithms with artificial agent controllers in electronic games,
called EvoMan, which is inspired in the action-platformer game Mega Man II. The
environment can be configured to run in different experiment modes, as single
evolution, coevolution and others. To demonstrate some challenges regarding the
proposed platform, as initial experiments we applied Neuroevolution
using Genetic Algorithms and the NEAT algorithm, in the context of competitively coevolving
two distinct agents in this game.~\footnote{This paper is a translation of
    \cite{karine2015}, published in Portuguese at Brazilian Congress on
Computational Intelligence, 2015.}

\end{abstract}


% \fi
\IEEEpeerreviewmaketitle

\section{Introduction}

The search for autonomous controllers is an important task of the Artificial
Intelligence field, aiming to create systems
able to take automatic decisions under uncertain environments. There are several
applications for such agents, varying from manufacturer industries to unmanned vehicles for exploring inhospitable places.

One issue to this field to be concerned about, is the lack of frameworks for assessing the performance
presented by learning algorithms in different scenarios. Testing environments    
which may help in such cases are those emulating electronic video games~\cite{thiha2009extending}~\cite{schrum2014evolving}. 
These testing environments allow to verify the capability of an algorithm when creating a
game playing agent that must succeed in the proposed game goal. Because of the
flexibility and variability of potential rules to an electronic game, it is
possible to simulate different levels of uncertainties encountered in real world
applications.

Specifically for the electronic gaming industry, this type of testing environment can help
during the try-out stage towards a game, in order to verify the difficulty and
feasibility to overcome a given challenge, or in order to develop a daunting AI
agent that meets the amusement level expected by a
game player~\cite{buchanan1991flow}.

Regarding the amusement level of a player, a game should not be too easy so that the
player is not daunted at all and neither too difficult so that the player
becomes frustrated by the game.

With the view to learn or improve their abilities, a human being may need to experience
the repetition of a determined situation many times until they are capable of
improving on the task at hand. After this first experience, if the person gets the chance to experience the situation again, the minimal knowledge and learning accumulated in the previous experience will be the only available. Thus, the person may react based on it, or ignore it and assume any other behavior, which makes it likely that it would be harder taking decisions in the earlier trials.

An agent controlled by some Artificial Intelligence on an electronic game
should evolve its behavior at a pace similar to a human player, so that the
challenge imposed is not unattainable by the player.

Considering these aspects, the proposed framework may contribute to practical
interests of the electronic games industry, which looks forward to creating
opponent characters capable of adapting to the profiles of human players in games,
providing a challenging but yet, satisfying experience. 

The goal of the present study is to introduce a new reference environment
called EvoMan, with the purpose of simulating and testing autonomous playing agents
in different game playing tasks. Formerly, in this paper we will illustrate how
this environment can be used as a testbed for a coevolutionary learning
experience with both autonomous players evolving to beat one another.

This paper is organized as follows: Section~\ref{sec:evoman} describes the
proposed environment, its main parameters and some more details about the
simulation modes; Section~\ref{sec:neuro} revises concepts of Neuroevolution and
Coevolution, as well as a survey of some related work;
Section~\ref{sec:experiments} describes the experimental methodology and results
obtained throughout the tests; and, finally Section~\ref{sec:conclusion}
concludes, pointing at some future research directions.
 
 \section{Background}
 \label{sec:neuro}

 In this Section we are reviewing some core concepts regarding the algorithms used
 during the experiments.

 \subsection{Artificial Neural Networks}
 
An Artificial Neural Network~\cite{gardner1998artificial} (ANN) is a
computational model inspired by how the animal brain works. In the most basic
and traditional ANN, the computational flow starts with some input variables
pertaining to a pattern recognition task. These input variables are then
blended through linear combinations and sent to the next layer of neurons where
they can activate such neurons or not, depending on the output of some chosen activation
function.

An ANN is commonly organized in layers of neurons~\cite{gardner1998artificial},
with the most trivial case being a single layer connecting the input neurons
directly to the output neurons, or the non-trivial case when multiple layers
generate intermediate results by the combination of variables until emitting an
output. The definition of an appropriate topology for the ANN is relevant to the
success of a learning task.
 
When the task at hand is a supervised learning problem, having a
sequence of sample input-output from which to learn, the weights for the inner
linear combinations of the ANN can be estimated by Gradient Descent algorithms, the most well-known being the Backpropagation~\cite{solanki2013review}.
This algorithm tries to correct the output generated by the network by
propagating the squared error regarding the desired output.

This adjustment is performed repeatedly until the weights correction
stabilizes. This procedure leads to a local optimum in the search space, that
may suffice for many applications.

Besides being a local search method, this algorithm requires the precise
knowledge of the desired output for every sampled input. However, sometimes, this
requirement is unfeasible to attend. For instance, when the task is an automation process for which the outcome is
only known after a sequence of outputs from the ANN. In this situation, there is
no way to know what the correct output for an specific input is, since the
current output depends on the previous outputs.

In such situations, researchers often resort to the Neuroevolutionary
algorithms. These algorithms optimize the weights of ANNs through a
gradient-less optimization algorithm, usually an evolutionary computational
meta-heuristic, by solely relying on a \textit{fitness} function roughly
measuring the quality of an ANN after some sequence of inputs.

 \subsection{Genetic Algorithms}

A Genetic Algorithm (GA)~\cite{holland1992genetic} is a meta-heuristic that
applies the principles of natural selection observed and described by Darwin, to
find solutions for a diverse range of problems.

In GA, the variables for a problem are encoded as artificial genes vectorized
as a chromosome. These chromosomes should be capable of representing the search
space of the problem being solved. The GA starts with a random
population of chromosomes and then repeats the following procedures:
recombining pairs of chromosomes (crossover), perturbing some selected
chromosomes (mutation) and carrying out a probabilistic selection favoring the fittest
(selection).

During the crossover operation, new chromosomes are created by combining
pairs~\footnote{in some variations the recombination is performed using more
than two chromosomes} of chromosomes from the current population. This
procedure should be performed in such a way that the offspring inherits characteristics within the genotype of both parents.

The mutation operation chooses a certain amount (usually a small amount) of
genes from a chromosome and changes their values at random. This operator was
proposed intending to promote diversity within the population and
avoid premature convergence.

After these operations, the algorithm will have created an offspring of new
chromosomes that will be evaluated and combined with the current population.
Afterwards, a selection procedure is applied in order to maintain the population
with a fixed size. One way to perform such selection is through tournament,
where two or more chromosomes are chosen completely at random and the fittest of
them is included into the new population. This procedure is then repeated until
the new population presents the required size.

 \subsection{Neuroevolution} 
 
As mentioned on a previous Section, whenever it is not possible to apply the
standard optimization algorithms to adjust the weights of an ANN, these weights are
optimized using an evolutionary algorithm, like a GA. This methodology is called 
Neuroevolution (NE)~\cite{floreano2008neuroevolution}.

However, besides optimizing the weights of the connections, the topology of the
ANN can also exert high influence on its performance regarding the
problem being solved. As such, it is reasonable trying to evolve the optimum
topology for the network, thus maximizing the performance of the learning
task.

For this purpose, the Neuroevolution of Augmenting Topologies
(NEAT)~\cite{stanley2002evolving} was created and it will be described in the
next subsection. 
 
 \subsection{NEAT algorithm}	
 
The algorithm starts with a random population of ANNs composed by the same topology,
consisting of a single layer relating the inputs to the outputs, but having different random connection weights.

The chromosome representation for each network is composed by two different
types of genes: 

 \begin{itemize}
 	\item  Node Genes, that represent inputs and neurons; 
 	\item  Connection Genes, which represent valued connections (weights) between neurons. 
 \end{itemize}
 
 These genes hold a flag indicating whether they are currently active or not.
 
 After creating the initial population, the algorithm performs the
 following steps at every generation:

 \begin{itemize}
 	
 	\item Speciation: divides the population into species, based on the
        similarities of the genomes.
 	
 	\item Fitness sharing:	adjusts the fitnesses of all genomes with the
        purpose of penalizing groups of chromosomes near the same base of
        attraction in the search space.
 	
 	\item Offspring size calculation:  calculates the size of the offspring to be
        produced by each species, proportional to the fitness of its members.
 	
 	\item Crossover: combines genomes of different species.
 	
 	\item Mutation:	changes the structure of some genomes randomly, by adding
        new nodes, creating new connections or varying the weights of existing connections.

 \end{itemize}

 \subsection{NEAT on Computational Intelligence in Games} 
 
 Trying to create an agent controller to play the game
 \textit{Frogs},~\cite{anconadeveloping} devised some experiments applying NEAT. 
 Twelve sensors were used so that the agent could sense the game current state,
 with these sensors measuring proximity of objects around it. Each individual
 was tested three times in the game, and the final fitness was the average score. 
 The score was the proximity of the frog to the objective area.
 
The application of NEAT for driving a car in TORCS using minimal inputs was discussed in~\cite{ihrigtraining}. 
The implementation presented in the paper used sensors towards the relative
angles of the car to the central line of the lane, and the car speed.
 
Using NEAT,~\cite{thiha2009extending} tried and succeed in figuring out a good fitness function for a soccer player robot, which was expected to learn how to avoid opponents, carry a ball and kick it to the goal.
 
Another interesting application was made in~\cite{schrum2014evolving}, for the Ms.Pacman game
playing framework. The objective of the game was to eat every capsule on the
screen while escaping the Ghosts that chased PacMan. When PacMan eats an special
pill, the Ghosts become edible and render more points for the player. In this
paper the problem was treated as a Multi-objective optimization problem, by
maximizing the score of the game while minimizing the number of neurons in the
hidden layer.
 
\subsection{Competitive Coevolution}
 
In nature, coevolution happens when a given species strengthens the selective
pressure over another species, forcing it to adapt in order to survive. When reciprocal, this
process may generate an arms race as both species try to surpass each other
for survival~\cite{dawkins1979arms}.

Competitive Coevolution (CC)~\cite{rawal2010constructing} is a method of
applying Evolutionary Algorithms, so that instead of evolving a single
population, two populations are mutually evolved through competition.

\begin{figure}
 	\centering
 	\includegraphics[width=3.5in]{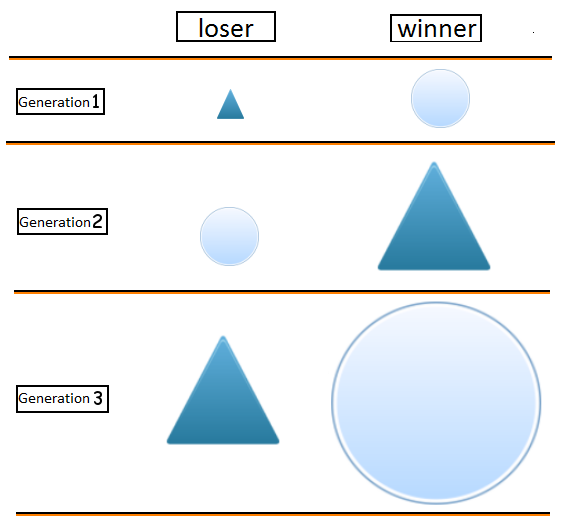}
 	\DeclareGraphicsExtensions.
 	\caption{Competitive Coevolution arms race: throughout generations we can see two geometric shapes (triangle and circle) competing, where the success criteria is having the largest size.}
 	\label{fig:coevo}
 \end{figure}

Figure~\ref{fig:coevo} illustrates an arms race scheme caused by a coevolution
process where, throughout the generations, the winner role is inverted between competing species, due to their adaptive changes. In this example, geometric shapes compete, considering the size as their fitness, so that the largest wins.
 
In games~\cite{cardona2013competitive}, the CC is sometimes used for the
predator-prey domain~\cite{pollack1997coevolution}. In this case, by evolving
synchronously to the population of the player, the population of enemies interferes 
in the environment, generating uncertainties on the objective space, thus
encouraging the exploration of the search space for a more general agent.

\subsection{Coevolution in Computational Intelligence for Games}
 
A Coevolutive algorithm was analyzed in~\cite{cardona2013competitive}, with the
Ms.Pacman framework. In this work it was verified that the coevolved controllers
achieved a better generalization for the game challenges than the standard
evolution. The authors also noticed it was harder evolving controllers for the
ghosts than for the Ms.Pacman, indicating that the success of this method may
depend on the problem domain.
 
In another interesting work, a robotic architecture was elaborated~\cite{gebeova2002experimental} aiming to evolve predator-prey behavior. 
The outcome allowed to observe that the evolved agent controllers acquired interesting conducts, as obstacle evasion, object discrimination and visual perception.

 \section{EvoMan framework}
 \label{sec:evoman}
 
The EvoMan~\footnote{https://github.com/karinemiras/evoman\_framework}
framework, proposed in~\cite{karine2015} is an environment for evolving game
playing agents for action-platformer games inspired by the classic Mega Man
II~\footnote{https://www.megaman.capcom.com}.

This framework contains eight different enemies against which the player agent must learn
how to beat, by performing one of the following simple actions: move left, move
right, jump, release jump and shoot.

It was developed in Python 2.7~\footnote{https://www.python.org} with the help of the library
Pygame~\footnote{http://www.pygame.org/}.

The game screen is composed of a rectangular area, that may contain some
obstacles depending on the stage. When the game starts, each character (the
player and the enemy) is positioned in opposing sides of the screen.

At every time step, the player and the enemy can perform one or more combined
actions to interact with the environment and their opponent, having the goal of
defeating them.

Both characters start the game with $100$ points of energy, which decrease
whenever they get hit by a projectile or hit each other. The character who
reaches $0$ points first loses the battle, making the other one winner.

The eight pre-programmed enemies perform their own distinct attacks, which are
stronger than the attack of the default player. They present a standard rule-based
behavior mimicking the original game.
 
\subsection{Simulation modes}
 
\begin{figure}[!t]
 	\centering
 	\includegraphics[width=3.6in]{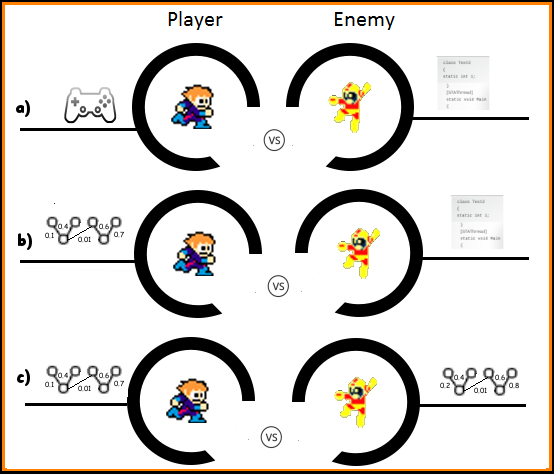}
 	\DeclareGraphicsExtensions.
 	\caption{Evoman simulation modes: examples of combinations.
 		a) player-character controlled by input device versus static enemy-character; b) player-character controlled by algorithm and static enemy-character; c) player-character controlled by algorithm versus enemy-character controlled by algorithm.
 	}
 	\label{fig:modes}
 \end{figure}

The framework allows running experiments using the combination of different simulation modes (Fig.~\ref{fig:modes}):

 \begin{itemize}
 	
 	\item \textbf{Human Player}: in which the player-character is controlled by
        some human input device (i.e., joystick). 
 	
 	\item \textbf{AI Player}: in which the player-character is controlled by a machine learning algorithm.
 	
 	\item \textbf{Static Enemy}: in which the enemy-character adopts a rule-based fixed attack/defence strategy, based on the original Mega Man II. 
 	
 	\item \textbf{AI Enemy}: in which the enemy-character is controlled by a machine learning algorithm. 
 	
 \end{itemize}
 
In this study we are using the simulation mode combinations ``AI Player VS Static
Enemy'' and ``AI Player VS AI Enemy'', to assess the learning of artificial
agent controllers for the characters of the game, using all the stages provided by the framework.
The second combination is the reproduction of a competitive coevolution process, to
verify the behavior of the learning agents.
 
\subsection{Game Sensors and Actions}
 
The environment provides $68$ variables about the game state, which act as sensors to the agents (AI player or AI enemy). The list below describes the sensors:

 \begin{enumerate} 
 	
 	\item Coordinates of the rectangles enveloping each character (8 sensors in total).  
 	\item Flag indicating weather each character is over some surface (2 sensors in total).
 	\item Time-steps left until the character is allowed to shoot a projectile
        again (2 sensors in total).
 	\item Flag indicating whether each character is shooting (2 sensors in total).
 	\item Vertical and horizontal acceleration for each character (4 sensors in total).
 	\item Direction each character is facing (2 sensors in total). 
 	\item Flag indicating whether each character is attacking or not (2 sensors in total). 
 	\item The coordinates of the rectangles enveloping each of the 3 projectiles of the player (12 sensors in total).  
 	\item The coordinates of the rectangles enveloping each of the 8 projectiles of the enemy (32 sensors in total).  
 	\item Flag indicating the enemy is immune to the player's attacks (1 sensor in total).   
 	\item Time-steps counter (1 sensor in total). 
 	
 \end{enumerate}
 
Not every sensor may be useful for the decision process made by the agent and also
some stages may require a smaller number of sensors (i.e., in some stages the
number of projectiles on the screen are limited).
 
 At every time step during a game run, after receiving the value for each
 sensor (game state) the player agent may perform up to $5$ actions:
 \textit{left}, \textit{right}, \textit{shoot}, \textit{jump} e
 \textit{release}. The \textit{release} action is used to mimic	the releasing of
 a joystick jumping button, interrupting the jumping process.

 The enemy agents perform additional actions whenever they present different attacks. These actions are labeled \textit{shootN} if $N$ ranging from $1$ to $6$.

\begin{figure}[!t]
 	\centering
 	\includegraphics[width=3.6in]{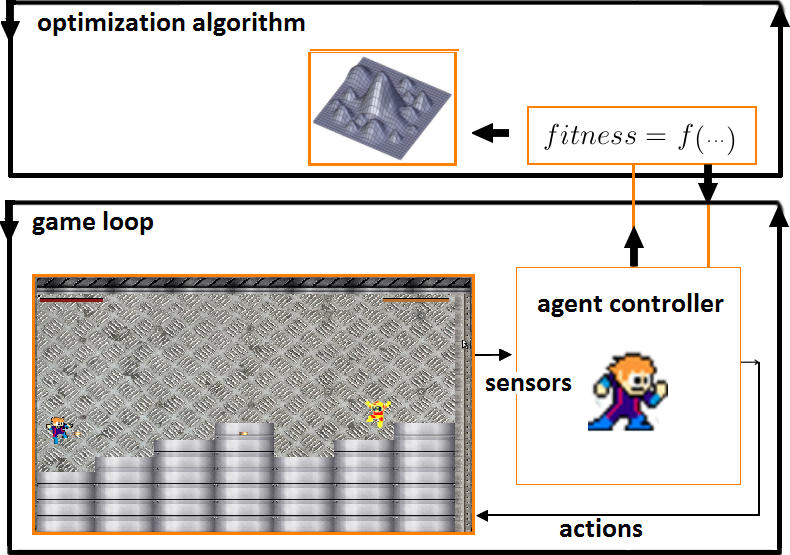}
 	\DeclareGraphicsExtensions.
 	\caption{In the framework, every time the user needs to test a solution (agent controller) using their  algorithm, they have to run a game loop, at which in each time-step sensors are provided to the agent being tested, as well as its decision actions are taken. The   
 		agent fitness evaluation is returned to the algorithm by the end of the loop run.
 	}
 	\label{fig:stream}
 \end{figure}

 The regular flow of an experiment using the framework is depicted by Figure~\ref{fig:stream}. 
 
 There are several parameters that might be supplied to the framework, in order to customize experiments. All of them are described in the documentation~\footnote{https://github.com/karinemiras/evoman\_framework} of EvoMan  framework.  
 
 \section{Experiments}
 \label{sec:experiments}
 
This section describes the experimental methodology performed in this paper
 regarding the coevolutionary approach, including a brief discussion about it.

 \subsection{Methodology}
 \label{sec:results}
 
Before assessing the behavior of a coevolutionary competition, we have performed
evolutions using the mode ``AI Player Vs Static Enemy`` in order to verify the
capability of each algorithm beating every enemy.

Afterwards, we experimented with the mode ``AI Player Vs AI Enemy``, in which
every agent (player or enemy) had a limited number of iterations to evolve a
better strategy than their opponent.
 
The experiments were performed using an ANN with a single layer having
its weights evolved by a standard Genetic Algorithm, and also an ANN evolved applying the NEAT algorithm.

The ANNs received each one of the $68$ sensors, normalized between $-1$ and $1$
as input, and the networks output $5$ neurons corresponding to one of each
possible action the agent could perform. The activation function used was the
Logistic Function and any output value higher than $0.5$ translated into
performing the corresponding action.

\subsubsection{Fitness function}
 
Intending to quantify the performance of each generated ANN, we established a
fitness function encompassing the criteria that we are willing to evolve. 
 
The proposed fitness function tries to balance the minimization in the energy of
the adversary, the maximization in the energy of the current player and the
minimization in the time taken to end the game. The fitness was modeled as
described by Eq.~\ref{eq:fitness}.
 
 \begin{equation}
 \label{eq:fitness}
 fitness = (100-e)^{\gamma} - (100-p)^{\beta} - ((\sum\limits_{i=1}^{\mbox{}(\vec{t})}100-p_i)/t)^{\alpha},
 \end{equation} 
 
 \noindent where \textit{e}, \textit{p} are the energy measures for the enemy
 and the player, respectively, in the range of $[0, 100]$; \textit{t}
 is the amount of time steps the game lasted. The constants  $\gamma$, $\beta$
 and $\alpha$ are weights to balance the importance of each term.
 
Experimentally, the optimal values found by these weights were $1$, $2$ and $2$
when evolving the main player, and $1$, $2$ and $3$ when evolving the enemy.
 
In order to promote a generalized behavior, capable of coping with a diverse
set of enemies, the final fitness for an individual is, the average value of the
fitness values obtained by fighting against the best individual of the opposite
population and also $4$ more randomly selected individuals from it.
 
\subsubsection{Training methods}
 
In order to test the evolution of artificial agents,  we applied an ANN having its weights
evolved by a Genetic Algorithm and an ANN having its weights and topology evolved by the NEAT algorithm.
 
At first, in order to generate a baseline, we evolved the player agent
against each enemy, separately, fixing the enemy behavior with an heuristic
approach.

Next, we performed experimentations of Co-Evolution between the main player
against the enemies. In this set of experiments, we alternated the
evolutionary process of each population by three generations per turn. Also,
due to time constraints, we tested only the NEAT algorithm.
 
The final energy measure for each agent by the
end of a fight for both agents are plotted in order to
perceive the co-evolutionary behavior, as they surpasses each other.

\subsection{Results}
\label{sec:results2}
 
Table~\ref{tab:indres} shows the average final energy obtained by each agent
evolved with GA and NEAT against the heuristic behavior of every enemy contained in the framework,
along with the p-values obtained by a t-Test.

 \begin{table}[!t]  
 	\caption{Mean energy of the main player obtained in the final generation
by Genetic Algorithm and NEAT.}
 	\label{tab:indres}
 	\centering

 	\begin{tabular}{|c||c||c||c||c||c|}
 		\hline
 		
 		\textbf{Enemy (game)} & \textbf{GA}   & \textbf{NEAT}  & \textbf{t-Test}  \\
 		
 		\hline FlashMan &	0	& 92 & \textit{p}  \textless 0.01 \\
 		\hline AirMan &	87 &	92 &  \textit{p}  \textless 0.01 \\
 		\hline WoodMan &	73 &	100 & \textit{p}  \textless 0.01  \\
 		\hline HeatMan &	48 &	84 & \textit{p}  \textless 0.01  \\
 		\hline MetalMan &	97 &	99 & \textit{p}  \textless 0.11 \\
 		\hline CrashMan &	87 &	54 & \textit{p}  \textless 0.01 	\\
 		\hline BubbleMan &	63 &	67 & \textit{p}  \textless 0.01 \\
 		\hline QuickMan &	77 &	88 &  \textit{p}  \textless 0.01 \\
 		
 		\hline

 	\end{tabular}
 \end{table}
 
As we see from this Table, every enemy can be beaten by both algorithms, being FlashMan the only exception, which only could be beaten by NEAT. Also, when
analyzing the remaining energy measures of the player, we notice that some enemies are
more difficult to beat than others. 

We can expect the co-evolutionary process to converge to an alternation
in which agents beat each other for  AirMan, MetalMan, BubbleMan and QuickMan when
using NEAT algorithm.  However, this same algorithm is expected to render unstable
results when confronting the other enemies.

In order to verify such assumption, we performed experiments and
reported the remaining energy obtained by the main player versus each enemy
with the pre-programmed heuristic (previous experiment) and against the
co-evolved enemy.

In Table~\ref{tab:cores} we can observe that there does not seem to be a correlation
between winning against the heuristic enemy and how easy it is to co-evolve
against it.

This Table shows the only beatable enemies by
co-evolution are AirMan, MetalMan, BubbleMan and QuickMan. Though WoodMan
was beatable during the co-evolution, it was by a small margin. 

It is worth considering that the enemies within this game have the advantage of
possessing more complex forms of attack which may hit the main player more
easily.
 
 \begin{table}[!t]  
 	\caption{Mean energy of the main player obtained in the final generation by
    NEAT with the Heuristic and Co-Evolutionbary experiments.}
 	\centering
 	
 	\begin{tabular}{|c||c||c||c||c|}
 		\hline
 		
 		\textbf{Enemy (game)}    & \textbf{Heuristic} & \textbf{Co-Evolution}  & \textbf{t-Test}  \\
 		
 		\hline FlashMan  	& 92 & 	0 & \textit{p}  \textless 0.01 \\
 		\hline AirMan   &	92 &	74 & \textit{p}  \textless 0.01 \\
 		\hline WoodMan  &	100 & 4 & \textit{p}  \textless 0.01 \\
 		\hline HeatMan   &	84 &	0 & \textit{p}  \textless 0.01 \\
 		\hline MetalMan   &	99 &	94 & \textit{p}  \textless 0.28 \\
 		\hline CrashMan   &	54 &	0  & \textit{p}  \textless 0.01 \\
 		\hline BubbleMan   &	67 &	22  & \textit{p}  \textless 0.01 \\
 		\hline QuickMan   &	88 &	70 & \textit{p}  \textless 0.01 \\
 		\hline 
 	\end{tabular}
 	\label{tab:cores}
 \end{table}
 
Nevertheless, the final analysis of the co-evolutionary systematization may not reflect the correct
performance towards implicate agents, since there is no guarantee that the point in time when the performance
was measured corresponds to the best performance achieved by the learning
algorithm.

\begin{figure}[!t]
 	\centering
 	\includegraphics[width=3.7in]{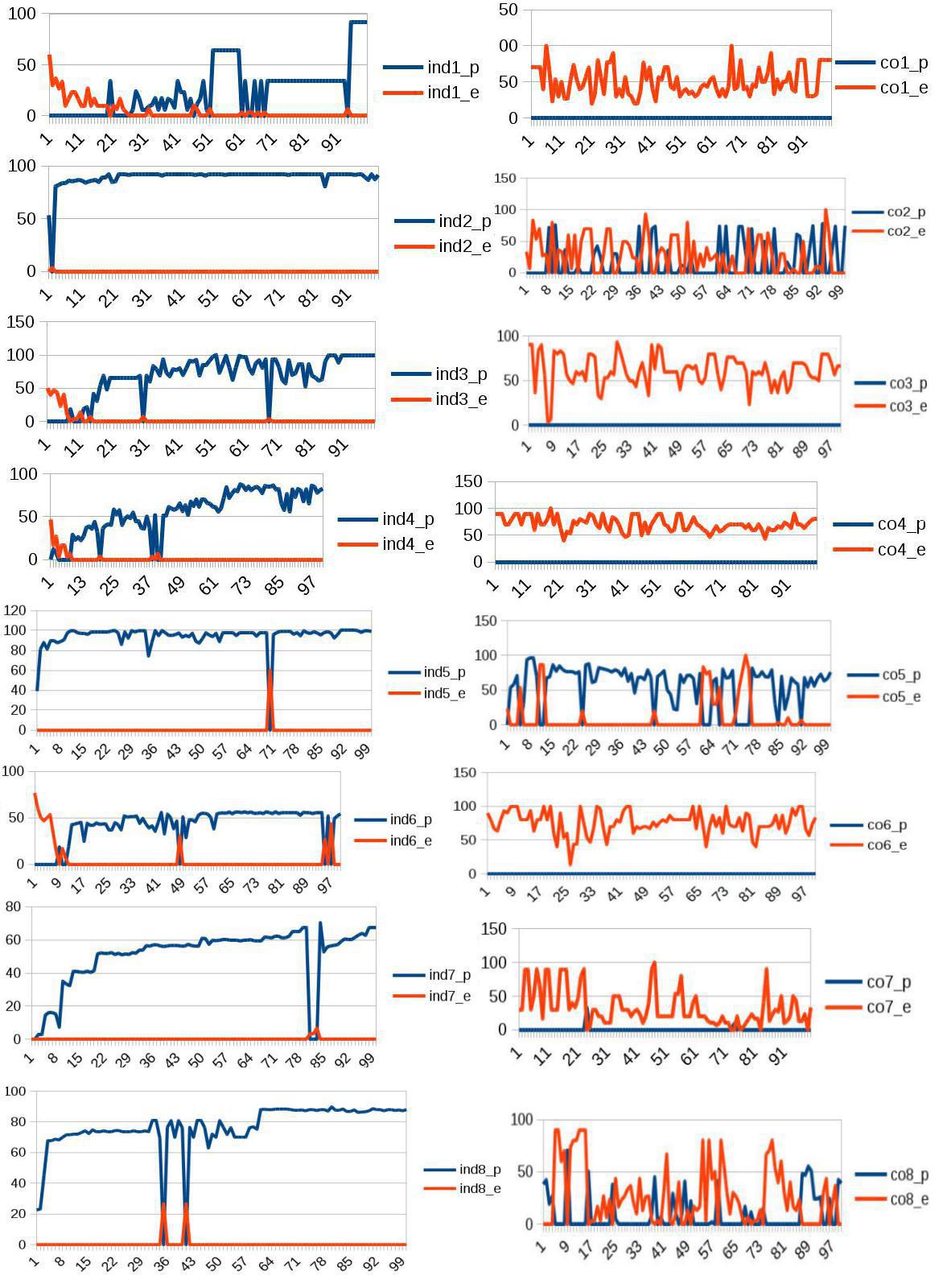}
 	\DeclareGraphicsExtensions.
    \caption{Final energy of the main player (\textit{p} suffix) and the enemy
        (\textit{e} suffix) throught $100$
    generations for the Heuristic experiments (\textit{ind} prefix) and
Co-Evolutionary experiments (\textit{co} prefix). The enemies are identified by
a number as: 1-FlashMan, 2-AirMan, 3-WoodMan, 4-HeatMan, 5-MetalMan, 6-CrashMan,
7-BubbleMan e 8-QuickMan.}
\label{fig:g}
 \end{figure}

In order to assess the overall behavior of the co-evolutionary approach, we have
plotted in Fig.~\ref{fig:g} the evolution regarding the energy of the main player (suffix \textit{\_p})
against the enemy (suffix \textit{\_e}) throughout the generations, for both 
Heuristic and Co-Evolutionary experimenting approaches.

From this Figure we learn that the easiest enemy to beat is AirMan, whereas	in the
Heuristic experiment the agent could learn how to win at an early generation. In
contrast, this is the one that shows the best arms race during the
Co-Evolutionary experiment, in which the winner agent alternates together with
the alternation of who is evolving.

On the other hand, we can notice that against FlashMan, the agent reached a good
result only at the later generations, meaning that it required a longer number
of generations to learn how to beat it. This reflects into the Co-Evolutionary
experiments, where we can see that the agent can learn how to attack it, but
never actually manages to beat it.

This behavior is similar to the one observed regarding WoodMan and HeatMan.

When playing against MetalMan, we can see that even though the agent can evolve
at early generations on the Heuristic experiment, the NEAT algorithm cannot
maintain the perfect solution as observed about AirMan. This is due to the
algorithm diversity control which uses a high mutation rate, and also the sensitivity
to change towards its chromosome representation. The CrashMan and BubbleMan enemies follow this same trend.

Finally, QuickMan presents a similar behavior to AirMan, for which the agent can
learn how to beat it at early generations and maintain this solution while in
Co-Evolutionary experiments it maintains an arms race against the enemy.

\section{Conclusion and future work}
\label{sec:conclusion}

This paper introduces the EvoMan game playing framework, which can be
used for testing learning algorithms capable of creating autonomous agents. The
purpose of such agents may be fighting against a pre-programmed opponent, by learning a general strategy to win against all opponents, or testing a co-evolutionary approach in
which the player and the enemy have both a limited number of generations to learn
how to beat each other, thus propelling an arms race.

In order to illustrate the capabilities of the proposed framework, we have performed two simple experiments:
i) applying two distinct learning algorithms against the pre-programmed behavior of each enemy,
ii) running a co-evolutionary learning experiment to verify if the player could keep up
with a constantly evolving enemy.

The heuristic experiments showed the feasibility of learning how to win against
the pre-programmed behavior of all enemies and, also, the difficulty level of each
enemy.

Regarding the co-evolutionary experiments, we learned that the main player
may require additional generations of advantage in order to generate an arms
race against some of the enemies.

Overall, this framework can help the Computational Intelligence in Games
community to test different tasks within a controlled environment, such as:
finding a general strategy able to defeat every opponent, a constrained
co-evolutionary approach in which the enemy behavior evolves in such a way that
its difficulty is always challenging against a human player.

For our future research we are going to investigate those mentioned tasks using this
framework. We are also going to improve the framework with different features such as
the possibility of acquiring the weapon from the enemy after the player has beaten it (similar to the original game), creating a combinatorial problem of which enemy to
beat first, and a continuous learning problem about how to use the
newly acquired weapons.


%\section*{Acknowledgment}



%\bibliographystyle{IEEEtran}

 \bibliographystyle{plain}
 \bibliography{evoman2}{}

\end{document}